\documentclass[letterpaper, 10 pt, conference]{ieeeconf}  

\IEEEoverridecommandlockouts                              

\usepackage{graphics} 
\usepackage{epsfig} 
\usepackage{mathptmx} 
\usepackage{times} 
\usepackage{amsmath} 
\usepackage{amssymb}  
\usepackage{cuted}
\usepackage{capt-of}
\usepackage{graphicx}
\usepackage{textcomp}
\usepackage{xcolor}
\usepackage{multirow}
\usepackage{booktabs}
\usepackage{arydshln}
\usepackage{color, colortbl}
\usepackage{microtype}
\usepackage[font=small, labelfont=bf]{caption}
\usepackage[colorlinks = true,
            linkcolor = black,
            urlcolor  = blue,
            citecolor = magenta,
            bookmarks = true, pagebackref]{hyperref}

\title{\LARGE \bf
Dur360BEV: A Real-world 360-degree Single Camera Dataset and Benchmark for Bird-Eye View Mapping in Autonomous Driving
}
\author{Wenke E\textsuperscript{1}, Chao Yuan\textsuperscript{1}, Li Li\textsuperscript{1}, Yixin Sun\textsuperscript{1}, Yona Falinie A. Gaus\textsuperscript{1}, \\ Amir Atapour-Abarghouei\textsuperscript{1}, Toby P. Breckon\textsuperscript{1, 2}\\
Department of \{\textsuperscript{1}Computer Science, \textsuperscript{2}Engineering\}, Durham University, UK
}

\begin{document}
\makeatletter
\let\@oldmaketitle\@maketitle%
\renewcommand{\@maketitle}{\@oldmaketitle%
    \centering
    \vspace*{1mm}
    \includegraphics[width=\linewidth]{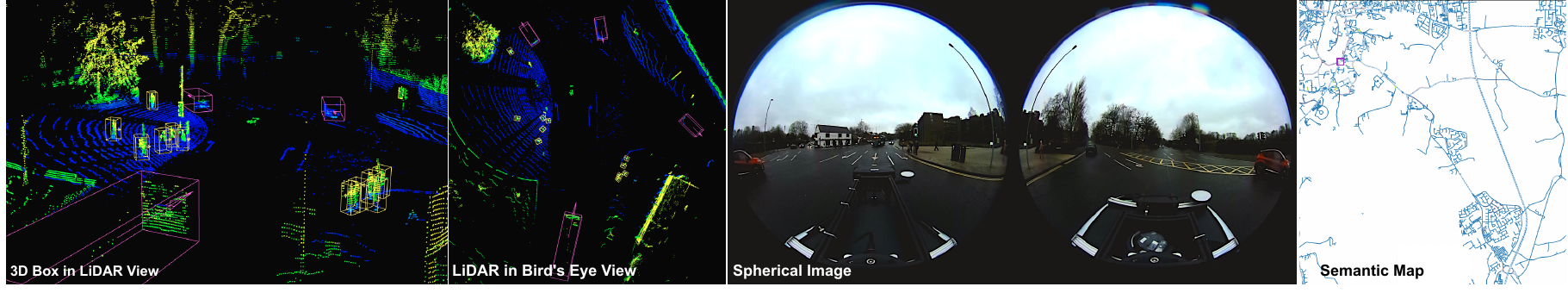}
    \captionof{figure}{An example from the Dur360BEV dataset where (from left-to-right) see 3D bounding box annotation for LiDAR, an exemplar LiDAR in Bird's Eye View (BEV), the dual-fisheye image from our spherical camera and our semantic map based on OpenStreetMap.}
    \label{fig:head}
}
\makeatother
\maketitle


\begin{abstract}

\textls[-4]{We present Dur360BEV, a novel spherical camera autonomous driving dataset equipped with a high-resolution 128-channel 3D LiDAR and a RTK-refined GNSS/INS system, along with a benchmark architecture designed to generate Bird-Eye-View (BEV) maps using only a single spherical camera. This dataset and benchmark address the challenges of BEV generation in autonomous driving, particularly by reducing hardware complexity through the use of a single 360-degree camera instead of multiple perspective cameras.
Within our benchmark architecture, we propose a novel spherical-image-to-BEV module that leverages spherical imagery and a refined sampling strategy to project features from 2D to 3D. Our approach also includes an innovative application of focal loss, specifically adapted to address the extreme class imbalance often encountered in BEV segmentation tasks, that demonstrates improved segmentation performance on the Dur360BEV dataset. The results show that our benchmark not only simplifies the sensor setup but also achieves competitive performance.}

\noindent
Code + Dataset: https://github.com/Tom-E-Durham/Dur360BEV
\end{abstract}
\section{INTRODUCTION} \label{s:intro}
\noindent
\textls[-5]{A spherical dual-fisheye camera, provides a full field of view (FoV) with dual fisheye lenses, capturing the entire environment in a single frame with just one device. This minimalist sensor setup offers a streamlined alternative to multi-camera systems. It is particularly well-suited for applications like autonomous driving, where a single spherical camera ensures full situational awareness while reducing hardware complexity, such as the need for multi-sensor calibration, synchronization, and connectivity~\cite{kitti2012,Waymo2020,nuscenes2020,Durlar2021}.}

While this setup simplifies the hardware, it introduces new challenges for image processing due to the significant radial distortion inherent in fisheye lenses. Previous studies~\cite{plaut20213d, rey2022360monodepth} have attempted to adapt pre-trained models designed for perspective images to fisheye images through techniques like rectification~\cite{xue2019learning} and data augmentation~\cite{duong2024robust}, but these methods often fall short in addressing the distortion. More recent work~\cite{li2022mode} demonstrates the effectiveness of convolutional neural networks (CNN) specifically designed for fisheye images, capturing detailed spatial information from a single spherical camera. However, utilizing this single-sensor data for generating accurate and reliable top-down views, such as bird’s-eye view (BEV) maps, remains an area requiring further exploration.





BEV maps are essential for autonomous driving, as they provide a unified top-down representation of the environment that helps in sensor fusion~\cite{liu2023bevfusion,man2023bev}, motion forecasting~\cite{Ettinger_2021_ICCV, zhang2024simpl}, and trajectory planning~\cite{zeng2019end,li2023trajectory}. These maps integrate raw sensor data into a format that is interpretable for downstream tasks, improving the system ability to predict vehicle motion~\cite{Wu_2020_CVPR, fiery2021} and plan paths~\cite{hu2023planning} effectively. Achieving accurate BEV maps requires the system to interpret spatial relationships within the scene and resolve issues related to distortion and occlusion, especially when relying solely on a single spherical camera.

While spherical imagery has been applied in previous research on tasks such as depth estimation~\cite{rey2022360monodepth,li2022mode,feng2022360}, the research direction of generating BEV maps using only a single spherical camera has not been thoroughly investigated. Most existing approaches~\cite{LSS2020,fiery2021,li2022bevformer, yang2023bevformer,PointBEV2024} depend on multiple-camera setups to mitigate challenges related to limited FoV. The reliance on additional sensors such as LiDAR~\cite{liang2022bevfusion} and radar~\cite{simplebev2023} further complicates system integration and increases cost, highlighting a gap for approaches that focus on minimising hardware (sensor) complexity and cost while maintaining overall perception performance.

We address this gap by proposing the first approach to generate BEV maps from a single spherical camera in the context of autonomous driving. To achieve this, we have collected the first autonomous driving dataset specifically featuring a single spherical camera image. Furthermore, we introduce a benchmark architecture that enables the generation of BEV maps using only one camera, providing a streamlined and efficient solution for autonomous driving applications.

\noindent
Overall, our contributions can be summarized as follows:
\begin{itemize}
\item[--] A novel large-scale real-world autonomous driving dataset comprising a (360\textdegree) spherical RGB camera, a high-fidelity 3D LiDAR (128 channels), and a GNSS/INS system. The first autonomous driving dataset with fully 3D bounding box annotation that features spherical camera modality.
\item[--] A benchmark for generating BEV maps from spherical images, with a novel spherical-image-to-BEV module that handles spherical distortions and maps 2D features onto a 3D sparse volume for accurate BEV representation.
\item[--] We introduce the use of focal loss, originally developed for object detection, as an innovative approach to address the extreme class imbalance in BEV segmentation. Our experiments demonstrate that this novel application of focal loss significantly improves segmentation performance, validating its effectiveness in the BEV domain.
\end{itemize}

\section{RELATED WORK} \label{s:related_work}
\noindent
We consider prior work in two related topic areas: autonomous driving datasets (Section \ref{ss:auto}) and vision based BEV model (Section \ref{ss:visbirdeye}).

\subsection{Autonomous Driving Datasets}
\label{ss:auto}
\begin{table*}[t]
  \centering
  \begin{tabular}{lcccccc}
    \toprule
    Dataset & Real/Synthetic & Frames & FPS & Camera & LiDAR & GPS  \\
    \midrule
    \rowcolor{gray!10}
    SynWoodScape~\cite{SynWoodScape2022} & Synthetic & 80K &10Hz& 4 fisheye cams & No & No \\
    
    OmniScape~\cite{omniscape2020}       & Synthetic & 10K & N/A & 2 fisheye cams  & No & No \\
    \rowcolor{gray!10}
    FB-SSEM~\cite{F2BEV2023}             & Synthetic & 20K &2Hz& 4 fisheye cams & No & No \\
    
    WoodScape~\cite{WoodScape2019}       & Real      & 10K &N/A& 4 fisheye cams & 64-channel & GPS only \\
    \rowcolor{gray!10}
    KITTI~\cite{kitti2012}               & Real      & 15K &10Hz& 1 stereo cam & 64-channel & GPS only \\
    
    KITTI-360~\cite{KITTI3602023}         & Real      & 78K &N/A& 1 stereo + 2 fisheye cams & 64-channel & GPS only \\

    \rowcolor{gray!10}
    Waymo~\cite{Waymo2020}               & Real      & 198K &10Hz& 5 perspective cams & 32-channel & GPS only \\
    
    nuScenes~\cite{nuscenes2020}         & Real      & 40K &1Hz& 5 perspective + 1 fisheye cams & 32-channel & GPS+RTK refined \\
    \rowcolor{gray!10}
    Lyft L5~\cite{lyft2021}              & Real      & 46 &1Hz& 7 perspective cams & 32-channel & GPS only \\
    DurLAR~\cite{Durlar2021}             & Real      & 0 & N/A & 1 stereo cam & 128-channel & GPS only \\
    \rowcolor{gray!10}
    \textbf{Dur360BEV (ours)}           & \textbf{Real}  &\textbf{32K}  &\textbf{10Hz}& \textbf{1 spherical cam}        &\textbf{128-channel} & \textbf{GPS+RTK refined}\\
    \bottomrule
  \end{tabular}
  \caption{Comparison between existing datasets (N.B. columns `Frames' and `FPS' in represent the frames labelled with 3D bounding box annotations and the frequency of these annotated frames in the dataset respectively; `N/A' means that the information is not provided or the dataset has no annotation).
    }
  \label{tab:dataset}
\end{table*}
\noindent
For autonomous driving, real-world datasets are crucial and numerous have been published in recent years.

\noindent
\textbf{Real dataset \textit{vs.} synthetic dataset.}
\textls[-5]{Autonomous driving datasets can typically be categorised into two types: real-world datasets~\cite{kitti2012,WoodScape2019,Waymo2020,nuscenes2020,lyft2021,Durlar2021,KITTI3602023} and synthetic datasets~\cite{omniscape2020,SynWoodScape2022,F2BEV2023}. Acquiring a real-world outdoor dataset requires considerable effort, including sensor setup, route planning, and data post-processing. Conversely, synthetic datasets generated from simulators offer comparable information with advantages such as time efficiency, cost savings and flexible data configurations-allowing for customized camera setups, precise 3D location data, and detailed annotation information ~\cite{dosovitskiy2017carla,airsim2017,gazebo2004}. However, they often lack the realism and unpredictability of real-world data, which can lead to gaps in model robustness and overfitting to specific characteristics of the synthetic environment ~\cite{play2017,train2018}. 
To address these issues, secondary solutions like domain adaptation techniques are often employed to bridge the gap and make models trained on synthetic data applicable to real-world scenarios ~\cite{daume2009, transfer2017, adaptation2019}.
Additionally, the absence of sensor noise and artifacts in synthetic datasets may result in models that struggle when applied to noisy real-world data~\cite{vkitti2016}.}

\noindent
\textbf{Perspective cameras \textit{vs.} fisheye cameras.}
Cameras are essential sensors for autonomous driving and in both real and synthetic datasets, different types of cameras can be configured. 
For example, datasets such as KITTI \cite{kitti2012}, Waymo \cite{Waymo2020}, and nuScenes \cite{nuscenes2020} predominantly use perspective (pin-hole) cameras to equip their vehicles. In nuScenes \cite{nuscenes2020}, five perspective cameras are used alongside a fisheye camera positioned at the rear of the vehicle. In contrast, datasets such as KITTI-360~\cite{KITTI3602023} and WoodScape \cite{WoodScape2019} rely exclusively on fisheye cameras, equipping their vehicles with multiple fisheye lenses. Fisheye lenses offer a significant advantage due to their wide FoV, allowing for 360-degree horizontal coverage with fewer cameras. For example, nuScenes \cite{nuscenes2020} employs six cameras to achieve a 360-degree visual coverage, whereas WoodScape \cite{WoodScape2019} requires four fisheye cameras, and KITTI-360 \cite{KITTI3602023} manages with only two. However, no dataset to date has utilised a single spherical camera in place of all other cameras and achieved comprehensive 360-degree visual information specifically targeting BEV recovery.






\subsection{Vision-based BEV Approaches}
\label{ss:visbirdeye}
\noindent
In autonomous driving, BEV map is useful for tasks such as object detection, path planning, and scene understanding. This process typically involves using multiple cameras positioned around the vehicle to capture the environment from various angles. The primary challenge is transforming 2D image data into a coherent 3D representation that can be accurately projected onto a BEV map. To address this challenge, several methods have been developed to lift 2D image features into 3D space before projecting them into a BEV map.

\subsubsection{Multiple Perspective Camera Models}
LSS \cite{LSS2020} pioneered the concept of lifting 2D features into 3D space before splatting them into a BEV map. This method laid the foundation for subsequent advancements in BEV generation. FIERY \cite{fiery2021} expanded upon this approach by integrating a multi-task framework, which uses uncertainty weighting to balance three critical sub-tasks: centerness, segmentation, and offset. SimpleBEV \cite{simplebev2023} further optimized the lifting strategy by introducing a bilinear-subsampling technique, which replaces the need for predicting depth distribution. PointBEV \cite{PointBEV2024} improved upon this process by employing a coarse-to-fine mechanism, which reduces the indexing size during the lifting phase. These methods all rely on input from six surrounding-view cameras to generate feature maps, leading to more accurate and efficient BEV representations for downstream applications in autonomous driving.

\subsubsection{Multiple Fisheye Camera Models}
While the majority of BEV models have focused on perspective camera setups, there has been growing interest in leveraging fisheye cameras due to their wide field of view. F2BEV \cite{F2BEV2023} is one of the few models that specifically addresses the challenges of generating BEV maps from multiple fisheye cameras. This approach is particularly advantageous for capturing a 360-degree view with fewer cameras, though it introduces additional complexities in handling the severe radial distortion inherent to fisheye lenses.

\section{DUR360BEV DATASET} \label{s:dataset}
\noindent
\textls[-5]{We introduce the first ever dataset that aims to use a single camera to solve real-world tasks in autonomous driving. Our dataset represents a shift towards the future of autonomous driving, where traditional systems often rely on an array of sensors, such as multiple cameras, LiDAR, and radar, working in tandem. By focusing on a single 360-degree camera, our approach not only simplifies the hardware setup but also reduces the complexity, cost, and power consumption of on-vehicle perception systems. Our Dur360BEV dataset comprises:}

\noindent
\textbullet\ \textbf{HD 360-degree camera imagery} in raw dual-fisheye format, where each pair of fisheye images are calibrated and can be in either equirectangular or cubemap formats.

\noindent
\textbullet\ \textbf{Annotated dense LiDAR pointclouds} which has the resolution of 128$\times$2048 and 3D bounding box annotation for vehicle, pedestrian and bicycles.

\noindent
\textbullet\ \textbf{RTK-corrected GNSS/INS positioning} delivering exceptional accuracy, providing at most centimeter-level position data and high-precision vehicle attitude measurements, with 0.03° accuracy in pitch/roll and 0.15° in slip angle, ensuring not only highly reliable vehicle localization but also precise self-attitude assessment.

\noindent
\textbullet\ \textbf{A High-Detail Semantic Map}, constructed using OpenStreetMap (OSM) in a geospatial database format, providing detailed environmental information surrounding the ego vehicle, as illustrated in Figure \ref{fig:head}. The use of OSM ensures that the database remains flexible and up-to-date, benefiting from ongoing contributions by the OSM community. 


\noindent
\textbullet\ \textbf{Ground truth BEV segmentation map} which contains object and map tile information in the local environment around the ego vehicle.

A comparison between existing datasets is shown in Table \ref{tab:dataset}. 
Our Dur360BEV dataset has the highest resolution in terms of the LiDAR sensor, a relatively high annotated FPS and is the only autonomous dataset that provides single spherical camera images. 
\noindent
\begin{table}[t]
  \centering
  
  \begin{tabular}{p{1.4cm}p{6.2cm}}
    \toprule
    Sensor & Details\\
    \midrule
    Camera & Spherical dual-fisheye camera (i.e., 360-degree camera, model: Ricoh Theta S), dual 1/2.3" 12M CMOS sensor, RGB image, 15Hz capture frequency, 1280x640 resolution, auto exposure, JPEG compressed, factory calibrated.\\ 
    LiDAR & 
    Ouster OS1-128 LiDAR sensor, 128 channel as vertical resolution, 2048 horizontal resolution, 10Hz capture frequency, 360 degree HFOV, -21.2 to 21.2 degree VFOV, 120m range @ \(>50\%\) detection probability, 100m range @ \(>90\%\) detection probability, 0.3cm range resolution.\\
    GNSS/INS & 
    OxTS RT3000v3 global navigation satellite and inertial navigation system, 100Hz capture frequency, 0.03 pitch/roll accuracy, 0.15 slip angle accuracy, centimeter level accuracy (with RTK corrections received via NTRIP).
    \\
    \bottomrule
  \end{tabular}
  \caption{Sensor details in Dur360BEV.}
  \label{tab_sensor}
\end{table}

\subsection{Sensor Setup}
\begin{figure}
    \centering
    \includegraphics[width=8cm]{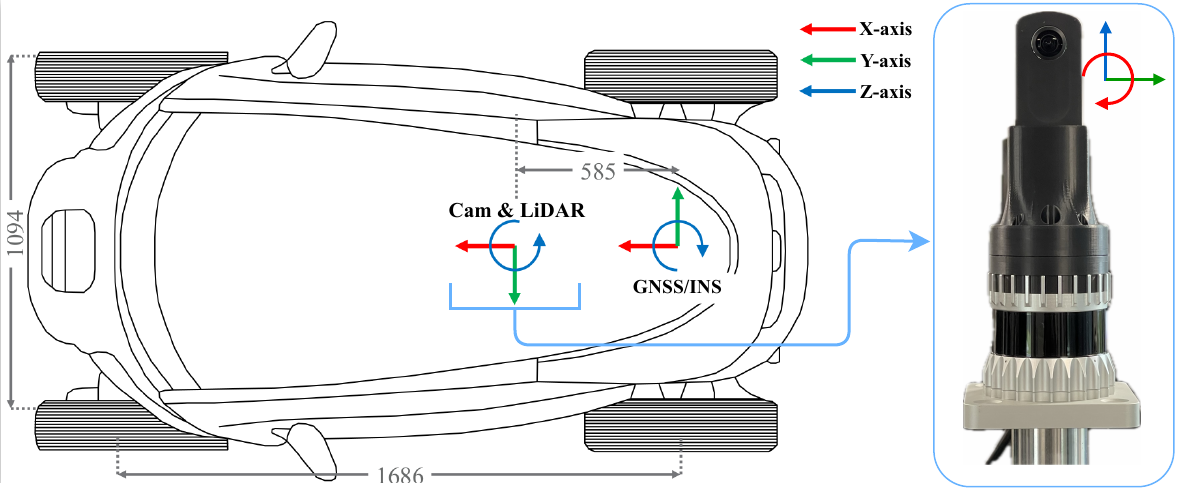}
    \caption{Sensor placement. Left: the top view of the vehicle equipped with sensors. Right: our spherical camera on top of the LiDAR. Both figures show the coordinates space for each sensor.}
    \label{fig:sensor_placement}
\end{figure}

\noindent
The dataset is collected by a spherical camera, a high-resolution LiDAR and a GNSS/INS navigation system calibrated and equipped on a Renault Twizy vehicle. The details of the sensor is shown in Table \ref{tab_sensor} and the setup is illustrated in Figure \ref{fig:sensor_placement}.

\subsection{Data Collection and Process}
\noindent
We collect data from various locations to ensure the dataset encompasses a diverse range of vehicles and traffic conditions. Specifically, we conduct data collection in four distinct areas of Durham, UK: the campus, highway, city center, and residential neighborhoods. These areas effectively represent the vast majority of driving environments across the UK. They include both straightforward scenarios, such as on highways where vehicle movements are relatively steady without parked cars on the roadside, and more challenging situations, such as in non-highway areas where vehicles might be parked in varying positions, and traffic patterns on the road become more complicated to predict due to additional traffic rules. 

\noindent
\textbf{Data synchronisation}: 
Given the inherent asynchrony in the data streams generated by different sensors operating at varied frequencies, we utilise the Robot Operating System (ROS Noetic) to achieve temporal alignment across different data sources based on their timestamps. The LiDAR, with its lower frame rate, serves as the reference sensor. We employ a synchronisation strategy that uses a queue size of 20 to handle incoming sensor messages and a slop parameter setting of 0.03 seconds to match messages from different sensors within this time window. This approach synchronises the dataset at 10 Hz, allowing for slight temporal discrepancies while ensuring accurate and coherent data integration, balancing the precision of alignment with the likelihood of successful message pairing.

\noindent
\textbf{3D bounding box annotations} were labeled using a combination of automated and manual processes on the Xtreme1 open-source annotation platform \cite{Xtreme1}. The high-resolution LiDAR sensor used in our setup facilitates both automated detection and manual labeling, as the dense point cloud data makes it easier for the model to detect objects and for human annotators to accurately label them. Initially, an integrated LiDAR object detection model was employed on the platform, successfully identifying approximately 60\% of the 3D bounding boxes for objects. Each bounding box annotation includes detailed data such as the 3D coordinates of the center, the rotation along the \textit{X}, \textit{Y}, and \textit{Z} axes, the size of the bounding box in three dimensions, the sensor distance, and the annotated point amount. Following the automated process, an experienced data annotator meticulously reviewed and manually annotated the remaining objects within a $100m \times 100m$ square area centered around the ego vehicle. The dataset is at a frequency of 10 Hz, and objects are labelled into three distinct classes: vehicle, pedestrian, and bicycle.

\subsection{Spherical Imagery}
\noindent
The proposed Dur360BEV dataset is the first to provide single spherical camera imagery specifically for autonomous driving tasks. Unlike previous datasets, such as nuScenes \cite{nuscenes2020} and Waymo \cite{Waymo2020}, which use multiple perspective cameras, or KITTI-360 \cite{KITTI3602023} and WoodScape \cite{WoodScape2019}, which rely on numerous fisheye cameras to capture 360-degree horizontal imagery, our approach utilizes a single spherical camera to achieve comprehensive coverage. This reduces both the number of sensors required and the input data size for models in this domain. As shown in \cite{simplebev2023}, while increasing input resolution in BEV model training on the nuScenes dataset can improve performance to some extent, it also significantly increases processing time, which is not ideal for real-time autonomous driving applications.

Our spherical imagery is crucial for calibrating the spherical camera with LiDAR data, enabling the accurate identification of corresponding pixels in an image based on the point cloud from the same frame. 
In practice, we find the best result by projecting 3D Cartesian coordinates \((X, Y, Z)\) onto spherical dual-fisheye image coordinates \((u, v)\) using a fourth-order polynomial transformation. First, the 3D coordinates are converted into spherical coordinates, where the azimuth angle \(\theta = \arctan2(Y, Z)\) and the polar angle \(\phi = \arctan\left(\frac{\sqrt{Y^2 + Z^2}}{X + \epsilon}\right)\) are calculated. The polar angle \(\phi\) is then mapped to a radius \(r(\phi)\) using the piecewise function:

\begin{equation}
r(\phi) = a_4\phi^4 + a_3\phi^3 + a_2\phi^2 + a_1\phi + a_0,
\end{equation}
where $a_0,a_1,a_2,a_3,a_4$ represent the coefficients of the polynomial that are determined through calibration to best fit the mapping between the spherical coordinates and the image plane.
\noindent
The 2D image coordinates
\(\mathbf{r} = \begin{pmatrix} x \\ y \end{pmatrix}\) 
are then computed as:
\begin{equation}
\mathbf{r} = r(\phi) \begin{pmatrix} \cos(\theta) \\ \sin(\theta) \end{pmatrix}.
\end{equation}
\noindent
Map the points from the front and back of the camera to their corresponding positions on the dual-fisheye image:
\begin{equation}
x = \begin{cases} 
    \frac{x + 1}{2}, & \text{if } X > 0, \\
    \frac{x - 1}{2}, & \text{if } X \leq 0. 
    \end{cases}
\end{equation}
\noindent
Let \( H \) and \( W \) denote the height and width of the image, respectively. The pixel coordinates \((u, v)\) are then given by the following expressions:
\noindent
\begin{equation}
u = \frac{x + 1}{2} \cdot W, \quad v = \frac{-y + 1}{2} \cdot H,
\end{equation}
avoiding any out-of-bounds coordinates:
\noindent
\begin{equation}
u_{\text{dist}} = \text{clip}(u, 0, W-1), \quad v_{\text{dist}} = \text{clip}(v, 0, H-1).
\end{equation}



\section{Methodology} \label{s:method}
\noindent
To leverage the advantages of the Dur360BEV dataset, we propose a novel benchmark task that takes spherical images as input to generate BEV map of the scene. Our benchmark architecture can be divided into two parts: Spherical-image-to-BEV module (Section \ref{ss:SI2BEV}) and multi-task framework with focal loss (Section \ref{ss:MTFLoss}).

\subsection{Spherical-Image-to-BEV module}
\label{ss:SI2BEV}
\noindent
As we replace the six input camera images used in previous work~\cite{LSS2020,fiery2021,li2022bevformer,yang2023bevformer,simplebev2023,PointBEV2024} with a single spherical camera to simplify the hardware setup and reduce redundancy, the conventional image-to-feature module is no longer applicable to our dataset. To address this, we introduce a novel application of the spherical-image-to-BEV module. This new module is specifically designed to handle the unique challenges posed by spherical imagery.

Building upon the foundational ideas in \cite{PointBEV2024}, our approach begins by feeding an RGB spherical image, with dimensions \(3 \times H \times W\), into a backbone network. This network outputs a feature map \(I \in \mathbb{R}^{C \times H \times W}\), where \(C, H, W \in \mathbb{N}\) represent the number of channels, height, and width of the feature map, respectively. Unlike traditional setups, which are tailored for perspective images from multiple cameras, our method is adapted to process the entire 360-degree field of view captured by a single spherical camera.

The backbone network extracts key features from the spherical image, which are then refined through a specifically tailored two-stage coarse-to-fine sampling strategy. This strategy is centered around what we call the Feature Pulling Process, which has been adapted to accommodate the distinct geometric properties of spherical images. By re-engineering the sampling geometry and refining the feature extraction process, we ensure that the module effectively captures and projects the spherical image features onto a BEV map, achieving accurate and reliable results.

\noindent
\textls[-5]{\textbf{The Feature Pulling Process} begins by taking a set of predefined 2D BEV points and generating pillars, each composed of 3D points with dimensions \(N_{\text{points}} \times 3\), where \(N_{\text{points}}\) represents the number of points sampled in this step. These 3D points are evenly spaced along the vertical axis in the BEV space. They are then projected onto the camera feature maps derived from the 360-degree imagery. Bilinear interpolation is applied to sample the corresponding 2D features, resulting in a high-dimensional feature volume with dimensions \(N_{\text{points}} \times C\). This feature volume is then processed by a decoder, such as a sparse U-Net, which compresses the features onto the 2D BEV plane, generating initial BEV predictions.}

\noindent
\textbf{Coarse sampling} applies the Feature Pulling Process to a broader set of 2D BEV points, generating a sparse 3D volume with dimensions \(N_{\text{coarse}} \times 3\). The resulting BEV predictions are used to identify high-confidence regions—those with the highest logit values—which are selected as anchor points. 

\noindent
\textbf{Fine sampling} then applies the Feature Pulling Process again based on the 3D points generated around the anchor points selected during the coarse sampling. This produces a refined feature volume with dimensions \(N_{\text{fine}} \times C\), which focuses on enhancing the representation of critical regions. The outputs from the fine stage are combined with those from the coarse stage to produce a final, densified BEV map.

The coarse-to-fine sampling strategy plays a crucial role in efficiently generating BEV maps by focusing computational resources on high-confidence regions in our spherical images, thereby potentially alleviating class imbalance issues to some extent. However, to further enhance the ability of the model to handle severe class imbalance, we integrate focal loss~\cite{ross2017focal} into our training process. 

\subsection{Multi-Task Framework and Loss Functions}
\label{ss:MTFLoss}
\noindent
\textls[-8]{Following the Multi-Task Framework~\cite{fiery2021}, our benchmark architecture incorporates three specialized segmentation heads—\textit{i.e.}, centerness, offset, and segmentation—each targeting a distinct aspect of the BEV map prediction. The centerness head predicts the likelihood of a location being the center of an object, the offset head estimates the spatial displacement from a predefined anchor point, and the segmentation head differentiates between foreground and background regions in the BEV map.}
For the centerness and offset tasks, we utilize a balanced mean squared error (MSE) loss and an absolute error loss, respectively. Given the significant class imbalance between foreground and background in BEV maps, we specifically apply focal loss~\cite{ross2017focal} to the segmentation head to enhance model focus on difficult-to-classify regions. 

Focal loss extends the standard cross-entropy (CE) loss, which is commonly used for binary classification. The cross-entropy loss is defined as $\text{CE}(p, y) = \text{CE}(p_t) = -\log(p_t)$, where $y \in {0,1}$ denotes the ground-truth label, and $p \in [0, 1]$ represents the predicted probability for the positive class ($y=1$). For uniformity, we define $p_t$ as:

\begin{equation} 
p_ {t}=\left\lbrace 
\begin{array}{ll}p, &\text{if $y = 1$}, \\ 
1 - p, &\text{otherwise.}\end{array}\right.
\end{equation}

One of the key challenges in training models for tasks like BEV segmentation is that standard cross-entropy loss tends to be dominated by straightforward examples, potentially overwhelming rare classes with small loss values. 
To better handle this, focal loss modifies the loss function by reducing the contribution of these simpler examples, thereby shifting the focus of training towards harder negatives. This is achieved by introducing a modulating factor $(1 - p_ {t})^\gamma$ with a tunable parameter $\gamma \geq 0$, leading to the focal loss formulation:
\noindent
\begin{equation} \mathrm {FL}(p_\mathrm {t}) = - (1 - p_\mathrm {t})^\gamma \log (p_\mathrm {t}). \end{equation}

This approach effectively reduces the influence of simpler classified examples, allowing the model to concentrate on learning from more challenging cases, which is crucial for handling imbalanced data in tasks like BEV segmentation.

\section{EXPERIMENTS} \label{s:exp}
\noindent
In this section, we outline the experimental setup used to evaluate our proposed SI2BEV module. We conduct a comparative analysis of two sampling strategies: dense grid sampling~\cite{simplebev2023} and a combination of sparse and dense sampling~\cite{PointBEV2024}, specifically focusing on their performance within our SI2BEV module on the Dur360BEV dataset. Additionally, we investigate the impact of varying the gamma parameter in the focal loss on BEV task performance.

\subsection{Experimental Setup} 
\noindent \textbf{Dataset:} 
Our experiments are conducted on the Dur360BEV dataset, a challenging real-world spherical image dataset tailored for autonomous driving applications. It consists of 16.4k point cloud frames, with 14.7k frames used for training and 1.6k for validation. The vehicle class is selected for training and evaluation. 

\noindent \textbf{Evaluation Protocol:} 
Ground truth BEV maps are generated using the 3D bounding box annotations for the vehicle class.  Pixels within these bounding boxes on the BEV map are labeled as positive, while all other pixels are labeled as negative. The evaluation metric is Intersection over Union (IoU), defined as the ratio of the overlap between predicted and ground truth positive regions to their union. Higher IoU values reflect better alignment and model performance.

\begin{figure}
    \centering
    \includegraphics[width=8cm]{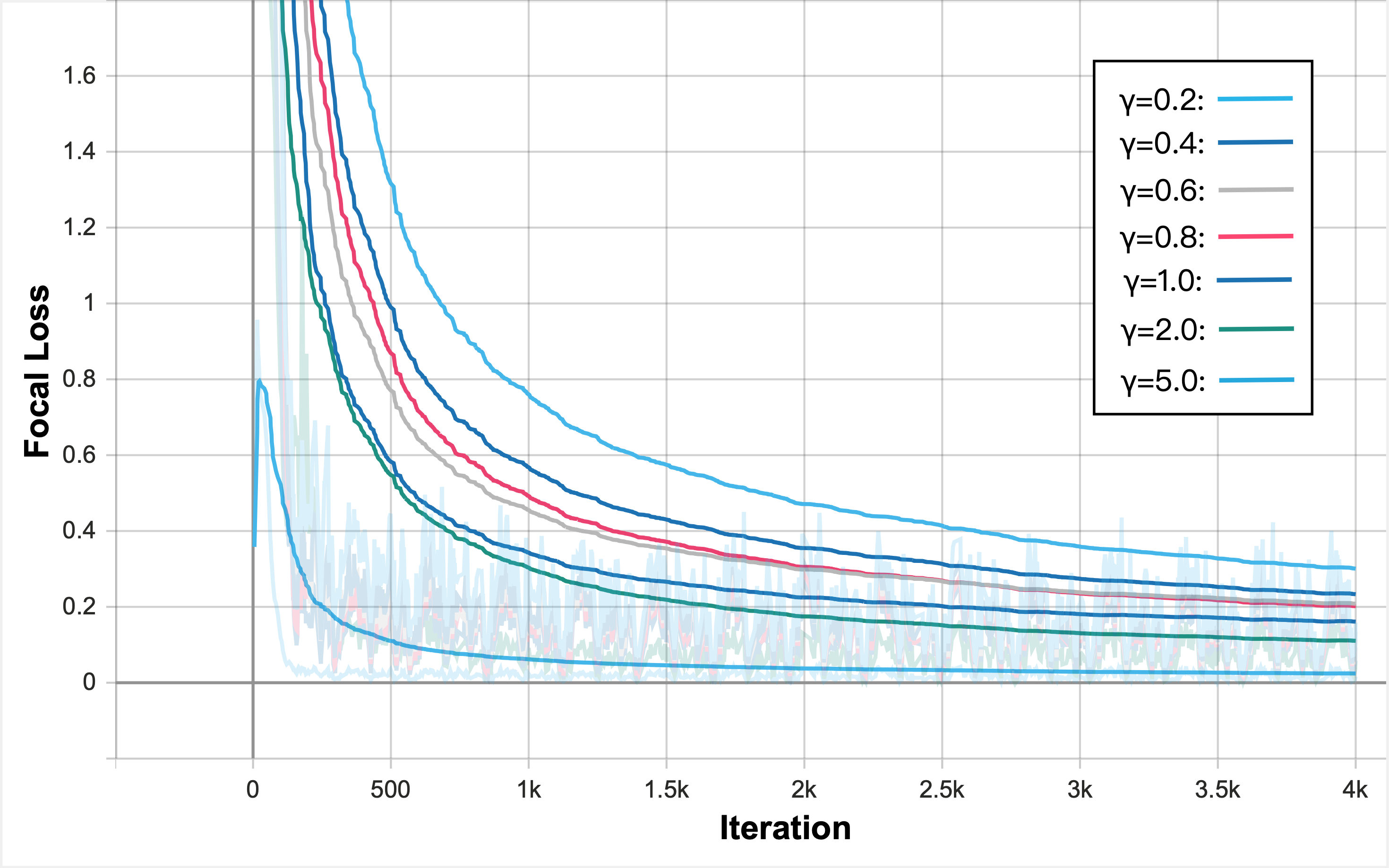}
    \caption{Validation loss curves for different values of $\gamma$. From top to bottom: $\gamma = 0.2, 0.4, 0.8, 0.6, 1, 2, 5$. The curves illustrate how the choice of $\gamma$ influences the convergence behavior during training.}
    \label{fig:stats_focal_loss_v}
    \vspace{-10pt}
\end{figure}

\noindent \textbf{Implementation Details:} 
For all experiments, the proposed architecture is trained over 4k iterations, each with 5 batches, using the AdamW optimizer~\cite{loshchilov2017decoupled}, with a learning rate of $\lambda=5e^{-5}$, weight decay $\mathrm{w}=10^{-7}$, and a 1-cycle learning rate schedule~\cite{smith2019super}. We closely monitor the validation loss throughout the training process, which consistently decreases to converge at a stable point, as shown in Figure \ref{fig:stats_focal_loss_v}. This convergence was achieved within the set 4000 iterations, beyond which the residuals of the loss function showed minimal change. The results confirm that the model effectively learns and stabilizes within this iteration limit, demonstrating the efficiency and reliability of our approach.

\subsection{Compare sampling strategies}
\noindent
We compare performance between the sparse/dense strategy proposed in our architecture and the dense grid sampling strategy used in SimpleBEV~\cite{simplebev2023}. Both methods are designed to detect objects within a 100m$\times$100m grid with a 50cm resolution resulting in a 200$\times$200 BEV map.

For a fair comparison, both sampling strategies were trained under identical conditions. The comparison results are summarized in Table \ref{tab:sample_result}. It is important to note that when $\gamma=0$, the focal loss simplifies to a standard Binary Cross Entropy (BCE) loss.

\begin{figure}[htb!]
    \centering
    \includegraphics[width=\linewidth]{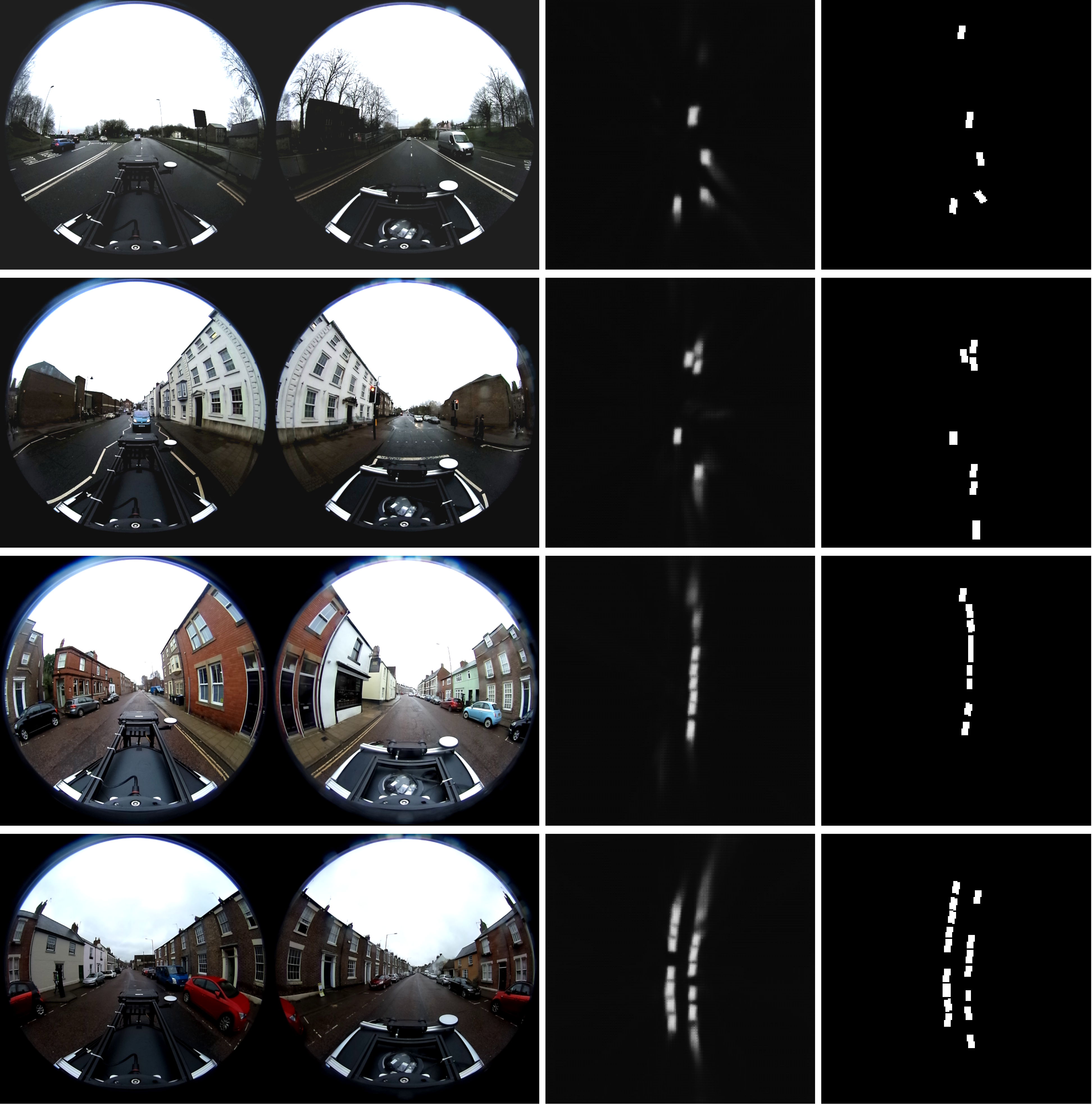}
    \caption{The inference visualisation of the Coarse/Fine sampling strategy and focal loss with $\gamma=2$ on Dur360BEV validation split. Left: Input image; Middle: Prediction; Right: Ground Truth Map.} 
    \label{fig:result_vis}
    \vspace{-10pt}
\end{figure}

As observed in Table \ref{tab:sample_result}, focal loss significantly enhances performance across both sampling strategies. The Dense Grid strategy shows a notable improvement in IoU by +1.6 at the 100m range, while the Coarse/Fine strategy achieves a +1.1 increase in IoU. These results highlight the effectiveness of focal loss in addressing class imbalance, particularly over extensive spatial ranges, thereby improving the accuracy of BEV segmentation. When considering the model complexity, the Coarse/Fine sampling strategy with a $\gamma=2$ setup demonstrates the best overall performance, balancing both IoU improvement and model efficiency, as shown in Table \ref{tab:model_complexity}. The qualitative visualisation of the result for this optimal setup is presented in Figure \ref{fig:result_vis}.

\begin{table}[t]
  \centering
  \addtolength{\tabcolsep}{-1pt}
  \begin{tabular}{lllllll}
    \toprule
    Strategy & $\gamma $ & Backbone & $\text{IoU}_{100}$ & $\text{IoU}_{50}$ & $\text{IoU}_{20}$ & Eff. Score\\
    \midrule
    \rowcolor{gray!10}
    Dense Grid  & 5 & RN-101 & 30.7 & 37.9 & 39.3 & 0.73\\
    Dense Grid  & 2 & RN-101 & 31.5 & 38.3 & 39.9 & 0.75\\
    \rowcolor{gray!10}
    Dense Grid  & 1 & RN-101 & \textbf{32.7} & \textbf{40.4} & \textbf{42.0} & 0.78\\
    Dense Grid  & 0 & RN-101 & 31.1 & 37.0 & 38.5 & 0.74\\
    \rowcolor{gray!10}
    Coarse/Fine & 5 & EN-b4 & 26.9 & 34.3 & 36.8 & 3.20\\
    Coarse/Fine & 2 & EN-b4 & \textbf{32.6} & \textbf{40.3} & \textbf{41.6} & \textbf{3.88}\\
    \rowcolor{gray!10}
    Coarse/Fine & 1 & EN-b4 & 28.9 & 36.3 & 39.4 & 3.44\\
    Coarse/Fine & 0 & EN-b4 & 31.5  & 38.9 & 39.7 & 3.75\\
    \rowcolor{gray!10}
    Coarse/Fine & 0.8 & EN-b4 & 29.5 & 36.7 & 37.6 & 3.51\\
    Coarse/Fine & 0.6 & EN-b4 & 27.5 & 35.2 & 37.4 & 3.27\\
    \rowcolor{gray!10}
    Coarse/Fine & 0.4 & EN-b4 & 31.0 & 38.3 & 40.0 & 3.69\\
    Coarse/Fine & 0.2 & EN-b4 & 31.0 & 38.4 & 40.3 & 3.69\\
    \bottomrule
  \end{tabular}
 \addtolength{\tabcolsep}{1pt}
  \caption{Comparison the BEV vehicle segmentation on Dur360BEV dataset between two sampling strategies. Computed on the validation split at different values of  $\gamma$ parameter in focal loss. ‘EN-b4’ and ‘RN-101’ stand for EfficientNet-b4~\cite{tan2019efficientnet} and ResNet101~\cite{he2016deep} respectively. $\text{IoU}_{100}, \text{IoU}_{50}, \text{IoU}_{20}$ represent the $\text{IoU}$ scores for the BEV maps in range 100m, 50m and 20m respectively. The ‘Eff. Score’ represents the ratio of the $\text{IoU}_{100}$ to the number of parameters (in millions). 
  All the model are trained on Dur360BEV training split, with batch size=6, learning rate=5e-5 and iterations=4000 for fairness.}
  \label{tab:sample_result}
\end{table}
\noindent

\begin{table}[htb!]
    \centering
    \setlength{\tabcolsep}{10pt} 
    \resizebox{\columnwidth}{!}{ 
    \begin{tabular}{llll}
    \toprule
        Strategy & C & $\text{IoU}_{100}$ & C:$\text{IoU}_{100}$   \\
        \midrule
        \rowcolor{gray!10}
        Dense Grid ($\gamma=1$) & 42.04 & 32.7 & 0.777 \\
        Coarse/Fine ($\gamma=2$) & 8.40 & 32.6 & 3.881 \\
        \bottomrule
    \end{tabular}
    }
    \caption{The model complexity is calculated by the ratio of the IoU to the number of model parameters (C) in millions.}
    \label{tab:model_complexity}
\end{table}

\section{CONCLUSIONS} \label{s:conclusion}
\noindent
We introduce Dur360BEV, the first large-scale autonomous driving dataset to feature a spherical RGB camera, high-fidelity 128-channel 3D LiDAR, and fully 3D annotated bounding boxes. This dataset simplifies hardware complexity while maintaining rich environmental data, advancing the state-of-the-art in BEV recovery for autonomous driving.

We also develop a benchmark architecture with the Spherical-Image-to-BEV (SI2BEV) module, effectively addressing the challenges of spherical imagery to produce accurate BEV maps. Our experiments further demonstrate that the incorporation of focal loss significantly enhances BEV segmentation performance, particularly in addressing class imbalance inherent in 360-degree camera datasets. This underscores the importance of considering class imbalance when generating BEV maps, especially in challenging environments. Collectively, our contributions provide robust tools and methodologies that enhance the development of autonomous driving technologies using a simpler low-cost, low-power sensing option.

\newpage
\bibliographystyle{bib/IEEEtran.bst} 
\bibliography{bib/main}

\begin{thebibliography}{10}
\providecommand{\url}[1]{#1}
\csname url@rmstyle\endcsname
\providecommand{\newblock}{\relax}
\providecommand{\bibinfo}[2]{#2}
\providecommand\BIBentrySTDinterwordspacing{\spaceskip=0pt\relax}
\providecommand\BIBentryALTinterwordstretchfactor{4}
\providecommand\BIBentryALTinterwordspacing{\spaceskip=\fontdimen2\font plus
\BIBentryALTinterwordstretchfactor\fontdimen3\font minus \fontdimen4\font\relax}
\providecommand\BIBforeignlanguage[2]{{%
\expandafter\ifx\csname l@#1\endcsname\relax
\typeout{** WARNING: IEEEtran.bst: No hyphenation pattern has been}%
\typeout{** loaded for the language `#1'. Using the pattern for}%
\typeout{** the default language instead.}%
\else
\language=\csname l@#1\endcsname
\fi
#2}}

\bibitem{kitti2012}
A.~Geiger, P.~Lenz, and R.~Urtasun, ``Are we ready for autonomous driving? the kitti vision benchmark suite,'' in \emph{2012 IEEE conference on computer vision and pattern recognition}.\hskip 1em plus 0.5em minus 0.4em\relax IEEE, 2012, pp. 3354--3361.

\bibitem{Waymo2020}
P.~Sun, H.~Kretzschmar, X.~Dotiwalla, A.~Chouard, V.~Patnaik, P.~Tsui, J.~Guo, Y.~Zhou, Y.~Chai, B.~Caine, \emph{et~al.}, ``Scalability in perception for autonomous driving: Waymo open dataset,'' in \emph{Proceedings of the IEEE/CVF conference on computer vision and pattern recognition}, 2020, pp. 2446--2454.

\bibitem{nuscenes2020}
H.~Caesar, V.~Bankiti, A.~H. Lang, S.~Vora, V.~E. Liong, Q.~Xu, A.~Krishnan, Y.~Pan, G.~Baldan, and O.~Beijbom, ``nuscenes: A multimodal dataset for autonomous driving,'' in \emph{Proceedings of the IEEE/CVF conference on computer vision and pattern recognition}, 2020, pp. 11\,621--11\,631.

\bibitem{Durlar2021}
L.~Li, K.~N. Ismail, H.~P.~H. Shum, and T.~P. Breckon, ``{DurLAR}: A high-fidelity 128-channel {LiDAR} dataset with panoramic ambient and reflectivity imagery for multi-modal autonomous driving applications,'' in \emph{2021 International Conference on 3D Vision (3DV)}.\hskip 1em plus 0.5em minus 0.4em\relax {IEEE}, 2021, pp. 1227--1237.

\bibitem{plaut20213d}
E.~Plaut, E.~Ben~Yaacov, and B.~El~Shlomo, ``3d object detection from a single fisheye image without a single fisheye training image,'' in \emph{Proceedings of the IEEE/CVF Conference on Computer Vision and Pattern Recognition}, 2021, pp. 3659--3667.

\bibitem{rey2022360monodepth}
M.~Rey-Area, M.~Yuan, and C.~Richardt, ``360monodepth: High-resolution 360deg monocular depth estimation,'' in \emph{Proceedings of the IEEE/CVF Conference on Computer Vision and Pattern Recognition}, 2022, pp. 3762--3772.

\bibitem{xue2019learning}
Z.~Xue, N.~Xue, G.-S. Xia, and W.~Shen, ``Learning to calibrate straight lines for fisheye image rectification,'' in \emph{Proceedings of the IEEE/CVF Conference on Computer Vision and Pattern Recognition}, 2019, pp. 1643--1651.

\bibitem{duong2024robust}
V.~H. Duong, D.~Q. Nguyen, T.~Van~Luong, H.~Vu, and T.~C. Nguyen, ``Robust data augmentation and ensemble method for object detection in fisheye camera images,'' in \emph{Proceedings of the IEEE/CVF Conference on Computer Vision and Pattern Recognition}, 2024, pp. 7017--7026.

\bibitem{li2022mode}
M.~Li, X.~Jin, X.~Hu, J.~Dai, S.~Du, and Y.~Li, ``Mode: Multi-view omnidirectional depth estimation with 360$^\text{o}$ cameras,'' in \emph{European Conference on Computer Vision}.\hskip 1em plus 0.5em minus 0.4em\relax Springer, 2022, pp. 197--213.

\bibitem{liu2023bevfusion}
Z.~Liu, H.~Tang, A.~Amini, X.~Yang, H.~Mao, D.~L. Rus, and S.~Han, ``Bevfusion: Multi-task multi-sensor fusion with unified bird's-eye view representation,'' in \emph{2023 IEEE international conference on robotics and automation (ICRA)}.\hskip 1em plus 0.5em minus 0.4em\relax IEEE, 2023, pp. 2774--2781.

\bibitem{man2023bev}
Y.~Man, L.-Y. Gui, and Y.-X. Wang, ``Bev-guided multi-modality fusion for driving perception,'' in \emph{Proceedings of the IEEE/CVF Conference on Computer Vision and Pattern Recognition}, 2023, pp. 21\,960--21\,969.

\bibitem{Ettinger_2021_ICCV}
S.~Ettinger, S.~Cheng, B.~Caine, C.~Liu, H.~Zhao, S.~Pradhan, Y.~Chai, B.~Sapp, C.~R. Qi, Y.~Zhou, Z.~Yang, A.~Chouard, P.~Sun, J.~Ngiam, V.~Vasudevan, A.~McCauley, J.~Shlens, and D.~Anguelov, ``Large scale interactive motion forecasting for autonomous driving: The waymo open motion dataset,'' in \emph{Proceedings of the IEEE/CVF International Conference on Computer Vision (ICCV)}, October 2021, pp. 9710--9719.

\bibitem{zhang2024simpl}
L.~Zhang, P.~Li, S.~Liu, and S.~Shen, ``Simpl: A simple and efficient multi-agent motion prediction baseline for autonomous driving,'' \emph{IEEE Robotics and Automation Letters}, 2024.

\bibitem{zeng2019end}
W.~Zeng, W.~Luo, S.~Suo, A.~Sadat, B.~Yang, S.~Casas, and R.~Urtasun, ``End-to-end interpretable neural motion planner,'' in \emph{Proceedings of the IEEE/CVF Conference on Computer Vision and Pattern Recognition}, 2019, pp. 8660--8669.

\bibitem{li2023trajectory}
H.~Li, P.~Chen, G.~Yu, B.~Zhou, Y.~Li, and Y.~Liao, ``Trajectory planning for autonomous driving in unstructured scenarios based on deep learning and quadratic optimization,'' \emph{IEEE Transactions on Vehicular Technology}, 2023.

\bibitem{Wu_2020_CVPR}
P.~Wu, S.~Chen, and D.~N. Metaxas, ``Motionnet: Joint perception and motion prediction for autonomous driving based on bird's eye view maps,'' in \emph{Proceedings of the IEEE/CVF Conference on Computer Vision and Pattern Recognition (CVPR)}, June 2020.

\bibitem{fiery2021}
A.~Hu \emph{et~al.}, ``{FIERY: Future Instance Prediction in Bird’s-Eye View From Surround Monocular Cameras},'' in \emph{Proceedings of the IEEE/CVF International Conference on Computer Vision}, 2021, pp. 15\,273--15\,282, accessed: Jun. 23, 2024.

\bibitem{hu2023planning}
Y.~Hu, J.~Yang, L.~Chen, K.~Li, C.~Sima, X.~Zhu, S.~Chai, S.~Du, T.~Lin, W.~Wang, \emph{et~al.}, ``Planning-oriented autonomous driving,'' in \emph{Proceedings of the IEEE/CVF Conference on Computer Vision and Pattern Recognition}, 2023, pp. 17\,853--17\,862.

\bibitem{feng2022360}
Q.~Feng, H.~P. Shum, and S.~Morishima, ``360 depth estimation in the wild-the depth360 dataset and the segfuse network,'' in \emph{2022 IEEE Conference on Virtual Reality and 3D User Interfaces (VR)}.\hskip 1em plus 0.5em minus 0.4em\relax IEEE, 2022, pp. 664--673.

\bibitem{LSS2020}
J.~Philion and S.~Fidler, ``Lift, splat, shoot: Encoding images from arbitrary camera rigs by implicitly unprojecting to 3d,'' in \emph{Computer Vision--ECCV 2020: 16th European Conference, Glasgow, UK, August 23--28, 2020, Proceedings, Part XIV 16}.\hskip 1em plus 0.5em minus 0.4em\relax Springer, 2020, pp. 194--210.

\bibitem{li2022bevformer}
Z.~Li, W.~Wang, H.~Li, E.~Xie, C.~Sima, T.~Lu, Y.~Qiao, and J.~Dai, ``Bevformer: Learning bird’s-eye-view representation from multi-camera images via spatiotemporal transformers,'' in \emph{European conference on computer vision}.\hskip 1em plus 0.5em minus 0.4em\relax Springer, 2022, pp. 1--18.

\bibitem{yang2023bevformer}
C.~Yang, Y.~Chen, H.~Tian, C.~Tao, X.~Zhu, Z.~Zhang, G.~Huang, H.~Li, Y.~Qiao, L.~Lu, \emph{et~al.}, ``Bevformer v2: Adapting modern image backbones to bird's-eye-view recognition via perspective supervision,'' in \emph{Proceedings of the IEEE/CVF Conference on Computer Vision and Pattern Recognition}, 2023, pp. 17\,830--17\,839.

\bibitem{PointBEV2024}
L.~Chambon, E.~Zablocki, M.~Chen, F.~Bartoccioni, P.~Pérez, and M.~Cord, ``{PointBeV: A Sparse Approach for BeV Predictions},'' in \emph{Proceedings of the IEEE/CVF Conference on Computer Vision and Pattern Recognition}, 2024, pp. 15\,195--15\,204.

\bibitem{liang2022bevfusion}
T.~Liang, H.~Xie, K.~Yu, Z.~Xia, Z.~Lin, Y.~Wang, T.~Tang, B.~Wang, and Z.~Tang, ``Bevfusion: A simple and robust lidar-camera fusion framework,'' \emph{Advances in Neural Information Processing Systems}, vol.~35, pp. 10\,421--10\,434, 2022.

\bibitem{simplebev2023}
A.~W. Harley, Z.~Fang, J.~Li, R.~Ambrus, and K.~Fragkiadaki, ``{Simple-BEV: What Really Matters for Multi-Sensor BEV Perception?}'' in \emph{2023 IEEE International Conference on Robotics and Automation (ICRA)}, 2023, pp. 2759--2765.

\bibitem{SynWoodScape2022}
A.~R. Sekkat, Y.~Dupuis, V.~R. Kumar, H.~Rashed, S.~Yogamani, P.~Vasseur, and P.~Honeine, ``{{SynWoodScape}}: {{Synthetic Surround-View Fisheye Camera Dataset}} for {{Autonomous Driving}},'' \emph{IEEE Robotics and Automation Letters}, vol.~7, no.~3, pp. 8502--8509, July 2022.

\bibitem{omniscape2020}
A.~R. Sekkat, Y.~Dupuis, P.~Vasseur, and P.~Honeine, ``The omniscape dataset,'' in \emph{2020 IEEE International conference on robotics and automation (ICRA)}.\hskip 1em plus 0.5em minus 0.4em\relax IEEE, 2020, pp. 1603--1608.

\bibitem{F2BEV2023}
E.~U. Samani, F.~Tao, H.~R. Dasari, S.~Ding, and A.~G. Banerjee, ``{{F2BEV}}: {{Bird}}'s {{Eye View Generation}} from {{Surround-View Fisheye Camera Images}} for {{Automated Driving}},'' in \emph{2023 {{IEEE}}/{{RSJ International Conference}} on {{Intelligent Robots}} and {{Systems}} ({{IROS}})}, Oct. 2023, pp. 9367--9374.

\bibitem{WoodScape2019}
S.~Yogamani, C.~Hughes, J.~Horgan, G.~Sistu, S.~Chennupati, M.~Uricar, S.~Milz, M.~Simon, K.~Amende, C.~Witt, H.~Rashed, S.~Nayak, S.~Mansoor, P.~Varley, X.~Perrotton, D.~Odea, and P.~Perez, ``{{WoodScape}}: {{A Multi-Task}}, {{Multi-Camera Fisheye Dataset}} for {{Autonomous Driving}},'' in \emph{2019 {{IEEE}}/{{CVF International Conference}} on {{Computer Vision}} ({{ICCV}})}.\hskip 1em plus 0.5em minus 0.4em\relax IEEE, 2019, pp. 9307--9317.

\bibitem{KITTI3602023}
Y.~Liao, J.~Xie, and A.~Geiger, ``{{KITTI-360}}: {{A Novel Dataset}} and {{Benchmarks}} for {{Urban Scene Understanding}} in {{2D}} and {{3D}},'' \emph{IEEE Transactions on Pattern Analysis and Machine Intelligence}, pp. 3292--3310, Mar. 2023.

\bibitem{lyft2021}
J.~Houston, G.~Zuidhof, L.~Bergamini, Y.~Ye, L.~Chen, A.~Jain, S.~Omari, V.~Iglovikov, and P.~Ondruska, ``One thousand and one hours: Self-driving motion prediction dataset,'' in \emph{Conference on Robot Learning}.\hskip 1em plus 0.5em minus 0.4em\relax PMLR, 2021, pp. 409--418.

\bibitem{dosovitskiy2017carla}
A.~Dosovitskiy, G.~Ros, F.~Codevilla, A.~Lopez, and V.~Koltun, ``Carla: An open urban driving simulator,'' in \emph{Conference on robot learning}.\hskip 1em plus 0.5em minus 0.4em\relax PMLR, 2017, pp. 1--16.

\bibitem{airsim2017}
S.~Shah, D.~Dey, C.~Lovett, and A.~Kapoor, ``Airsim: High-fidelity visual and physical simulation for autonomous vehicles,'' in \emph{Field and Service Robotics: Results of the 10th International Conference}.\hskip 1em plus 0.5em minus 0.4em\relax Springer, 2017, pp. 621--635.

\bibitem{gazebo2004}
N.~Koenig and A.~Howard, ``Design and use paradigms for gazebo, an open-source multi-robot simulator,'' in \emph{IEEE/RSJ International Conference on Intelligent Robots and Systems (IROS)}.\hskip 1em plus 0.5em minus 0.4em\relax IEEE, 2004, pp. 2149--2154.

\bibitem{play2017}
S.~R. Richter, V.~Vineet, S.~Roth, and V.~Koltun, ``{Playing for data: Ground truth from computer games},'' \emph{International Journal of Computer Vision}, vol. 125, no. 1-3, pp. 127--144, Dec. 2017.

\bibitem{train2018}
J.~Tremblay \emph{et~al.}, ``{Training deep networks with synthetic data: Bridging the reality gap by domain randomization},'' in \emph{Proceedings of the IEEE Conference on Computer Vision and Pattern Recognition Workshops (CVPRW)}, 2018, pp. 1082--1090.

\bibitem{daume2009}
H.~D. III, ``Frustratingly easy domain adaptation,'' in \emph{Proceedings of the 47th Annual Meeting of the Association for Computational Linguistics (ACL)}.\hskip 1em plus 0.5em minus 0.4em\relax Association for Computational Linguistics, 2009, pp. 256--263.

\bibitem{transfer2017}
J.~Hoffman, D.~Wang, F.~Yu, and T.~Darrell, ``Fcns in the wild: Pixel-level adversarial and constraint-based adaptation,'' in \emph{Proceedings of the IEEE/CVF Conference on Computer Vision and Pattern Recognition (CVPR)}.\hskip 1em plus 0.5em minus 0.4em\relax IEEE, 2017, pp. 3427--3436.

\bibitem{adaptation2019}
Y.~Wang, W.~Liu, and Z.~Wang, ``Unsupervised domain adaptation for autonomous driving with synthesized data,'' in \emph{Proceedings of the IEEE/CVF Conference on Computer Vision and Pattern Recognition (CVPR)}.\hskip 1em plus 0.5em minus 0.4em\relax IEEE, 2019, pp. 1238--1247.

\bibitem{vkitti2016}
A.~Gaidon, Q.~Wang, Y.~Cabon, and E.~Vig, ``{Virtual KITTI: Analyzing visual systems with synthetic scenes},'' in \emph{Proceedings of the IEEE Conference on Computer Vision and Pattern Recognition (CVPR)}, 2016, pp. 578--585.

\bibitem{Xtreme1}
\BIBentryALTinterwordspacing
L.~A. .~D. Foundation, ``Xtreme1 - the next gen platform for multisensory training data,'' 2023, software available from https://github.com/xtreme1-io/xtreme1/. [Online]. Available: \url{https://xtreme1.io/}
\BIBentrySTDinterwordspacing

\bibitem{ross2017focal}
T.-Y. Ross and G.~Doll{\'a}r, ``Focal loss for dense object detection,'' in \emph{proceedings of the IEEE conference on computer vision and pattern recognition}, 2017, pp. 2980--2988.

\bibitem{loshchilov2017decoupled}
I.~Loshchilov, ``Decoupled weight decay regularization,'' \emph{arXiv preprint arXiv:1711.05101}, 2017.

\bibitem{smith2019super}
L.~N. Smith and N.~Topin, ``Super-convergence: Very fast training of neural networks using large learning rates,'' in \emph{Artificial intelligence and machine learning for multi-domain operations applications}, vol. 11006.\hskip 1em plus 0.5em minus 0.4em\relax SPIE, 2019, pp. 369--386.

\bibitem{tan2019efficientnet}
M.~Tan, ``Efficientnet: Rethinking model scaling for convolutional neural networks,'' \emph{arXiv preprint arXiv:1905.11946}, 2019.

\bibitem{he2016deep}
K.~He, X.~Zhang, S.~Ren, and J.~Sun, ``Deep residual learning for image recognition,'' in \emph{Proceedings of the IEEE conference on computer vision and pattern recognition}, 2016, pp. 770--778.

\end{thebibliography}


\end{document}